\documentclass{article}

\usepackage{arxiv}

\usepackage[utf8]{inputenc} 
\usepackage[T1]{fontenc}    
\usepackage{hyperref}       
\usepackage{url}            
\usepackage{booktabs}       
\usepackage{amsfonts}       
\usepackage{nicefrac}       
\usepackage{microtype}      
\usepackage{graphicx}
\usepackage{doi}
\usepackage{amsmath, amssymb}

\title{\emph{TSCAN} : Dialog Structure discovery using SCAN}

\author{ {\hspace{1mm}Apurba Nath}\thanks{https://voicezen.ai} \\
	\texttt{apurba@voicezen.ai} \\
	\And
	{\hspace{1mm}Aayush Kubba} \\
	\texttt{aayush@voicezen.ai} \\
}

\renewcommand{\headeright}{TSCAN}
\renewcommand{\undertitle}{Adaptation of SCAN to text data}

\hypersetup{
pdftitle={TSCAN: Dialog Structure discovery using SCAN},
pdfauthor={Apurba, Aayush},
pdfkeywords={Dialog Structure Discovery, SCAN},
}

\begin{document}
\maketitle

\begin{abstract}
    Can we discover dialog structure by dividing utterances into labelled clusters. Can these labels be generated from the data. Typically for dialogs we need an ontology and use that to discover structure, however by using unsupervised classification and self-labelling we are able to intuit this structure without any labels or ontology. In this paper we apply SCAN (Semantic Clustering using Nearest Neighbors) to dialog data. We used BERT for pretext task and an adaptation of SCAN for clustering and self labeling. These clusters are used to identify transition probabilities and create the dialog structure. The self-labelling method used for SCAN makes these structures interpretable as every cluster has a label. As the approach is unsupervised, evaluation metrics is a challenge, we use statistical measures as proxies for structure quality. 

\end{abstract}


\section{Introduction and prior work}

\paragraph{}
Dialog structure discovery is an important problem given the increased efforts in automation of text and voice response systems. Unlike the simulated dialogs or human bot interactions, human human interactions are richer (larger vocab and variation) and have more number of turns. For example typical dialog datasets created from SimDial have less than 1.5k vocab spread with 20 n-grams covering majority of generated responses. Compared to this our internal human human task oriented dialog incldues over 5k of common closed (proper nouns excluded) vocab with the most common 20 n-grams failing to cover even 1\% of utterances. 

\paragraph{}
Many approaches rely on generational models which are trained on the dialog data e.g VRNN approach (Shi, Zhao) \cite{qiu-etal-2020-structured}  or DVAE-GNN (Xu, Che) \cite{xu2020DDG}. There are also approaches using transformers like BERT models. In our experience BERT when trained on large in-domain data captures semantic information abundantly, we can cluster on these embeddings but balance and interpretability of these clusters is still a challenge. 

\paragraph{}
On image classification without labels SCAN \cite{vangansbeke2020scan} has achieved great results. Their approach comprises of obtaining semantically meaningful features, learning a clustering approach and then self-labelling for interpretable clusters. They use image transforms and nearest neighbors in this work. They use confidence and consistency both as part of their objective function while training the clustering model which creates balanced clusters. It is also not negatively impacted by overclustering \footnote{in fact in our experiments we depend on overclustering}.

\section{Our approach}
\paragraph{}
Each dialog is made of T turns  $(A_1, U_1), (A_2, U_2), ..., (A_t, U_t)$ where $A_t$ is the agent utterance at t-th dialogue turn and $U_t$ the user utterance \footnote{Unlike other approaches we do not classify the user utterance as a response. In our datasets we have seen multiple cases where user utterance is a query and the agent utterance is the response}. These dialogs are task oriented  and may have multiple exchanges (multiple tasks) in the same dialog. Our goal is to eventually find out any correlation between $A_t$ and $U_t$ and $U_t$ and $A_{t+1}$. We try to first reduce the size of the space (because of vocab variety) by assiging the utterances to clusters. With a 20 state cluster, now this becomes a problem of matching the clusters among each other. For example assuming $A_t$ belongs to agent cluster $AC_0$ and user utterance $U_t$ belongs to $UC_3$ we can group them with any other turn which similarly have $AC_0$ and $UC_3$. We create transition probabilities between the cluster combinations $(AC_i, UC_j)$, these transition probabilities are then used for dialog states.

\paragraph{}
A simplified version of these steps are
\begin{itemize}
    \item Use in-domain trained BERT for semantic embeddings.
    \item Train SCAN model with nearest neighbors on 10k $A_*$ agent utterances and $U_*$ user utterances
    \item Create clusters using this model and apply self-labels on it
    \item For each Dialog turn assign the agent cluster and customer cluster 
    \item Create a transition map between agent and customer turn and customer to next agent turn
    \item Create dialog flows with these transition states, each cluster is represented by it's equivalent label
\end{itemize}

\subsection{Models:}
The bert model needs to satisfy the equation (1) of SCAN paper, replicated here for convenience

\begin{equation}
min_\theta d(\phi_\theta (X_i),\phi_\theta(T[X_i]))
\end{equation}

finetuning or training on large volume of in-domain data helps us create such a model. Any MLM evaluation task can be used to check the semantic quality of the model.

The SCAN model needs to satisy the equation (2) of the SCAN paper, a simlified form of that equation is
\begin{equation}
    loss = consistency\_loss - entropy\_weight * entropy\_loss
\end{equation}
consistency loss is BCE between anchors and neighbors while entropy loss is mean of anchors probability. The entropy constituent helps in balancing the distributions within the clusters.

\subsection{Our experiences}
Unlike original SCAN implementation we do not use transformations or augmentation,  instead we rely on the variety of data to provide the relevant neighbors. We also do not build a pretext model but use a BERT model trained on in-domain data for the same. Our experiments show that inspite of these deviations from the SCAN approach we are able to create a interpretable dialog structure from the balanced well defined clusters created by SCAN. We use two statistical measures as evaluation metrics to understand the cluster quality and our experiments show that TSCAN (text SCAN) does better than K-means on both these measures.

\subsection{Evalaution Metrics}
The goal of clustering is to evenly balance the utterances between the clusters. This means we should not have any cluster that is too big. To compute the distribution score, we use

\paragraph{Distribution}
We want the clusters to be balanced, that means each cluster should have nearly the same number of members. A good measure of the same is 
\begin{equation}
    \sum xlog(x)
\end{equation}
where x is the ratio of members vs total elements. Though this number is not comparable across cluster sizes, within a cluster size it is a good indicator of the distribtuion. For comparison across cluster sizes we can use deviation from ideal distribution.

\paragraph{Confidence}
We expect similar utterances to end up in the same cluster. As we already have some pre-trained intent models, we can check that utterances with the same intent end up together. We want the number of clusters to be as low as possible . A good measure of togetherness is the mean and standard deviation of cluster membership. For exmaple in case we have a greeting intent, we would want all the greeting intents to end up in the same cluster. A scenario where it is spread between 3 different clusters out of 20 is better than where it is spread between 8. Mean and standard deviation of these two scenarios give a good indication of the distribution.

\section{Results}
For internal dataset, we were able to arrive at interpretable dialog states.

\paragraph{Consistency}
For the clustering approach, for a 20 cluster SCAN vs K-means approach, K-means shows a distribution score of -2.64, SCAN achives -2.77 while an ideal distribution is -2.995

\paragraph{Confidence}
Similarly for two intents greeting and payment\_inquiry, the results were

\begin{verbatim}
    intent: payment_inquiry
    with K-means
    nobs=8, minmax=(6, 43), mean=12.12, variance=162.98, skewness=2.07, kurtosis=2.63
    
    with Scan
    nobs=6, minmax=(6, 60), mean=16.16, variance=468.97, skewness=1.73, kurtosis=1.07
    
    intent: greeting 
    with K-means
    nobs=6, minmax=(0, 91), mean=16.17, variance=1347.77, skewness=1.78, kurtosis=1.18
    
    with Scan
    nobs=5, minmax=(0, 95), mean=19.4, variance=1786.30, skewness=1.50, kurtosis=0.25
\end{verbatim}

\begin{figure}[!htb]
\centering
\includegraphics[width=6cm, height=6cm]{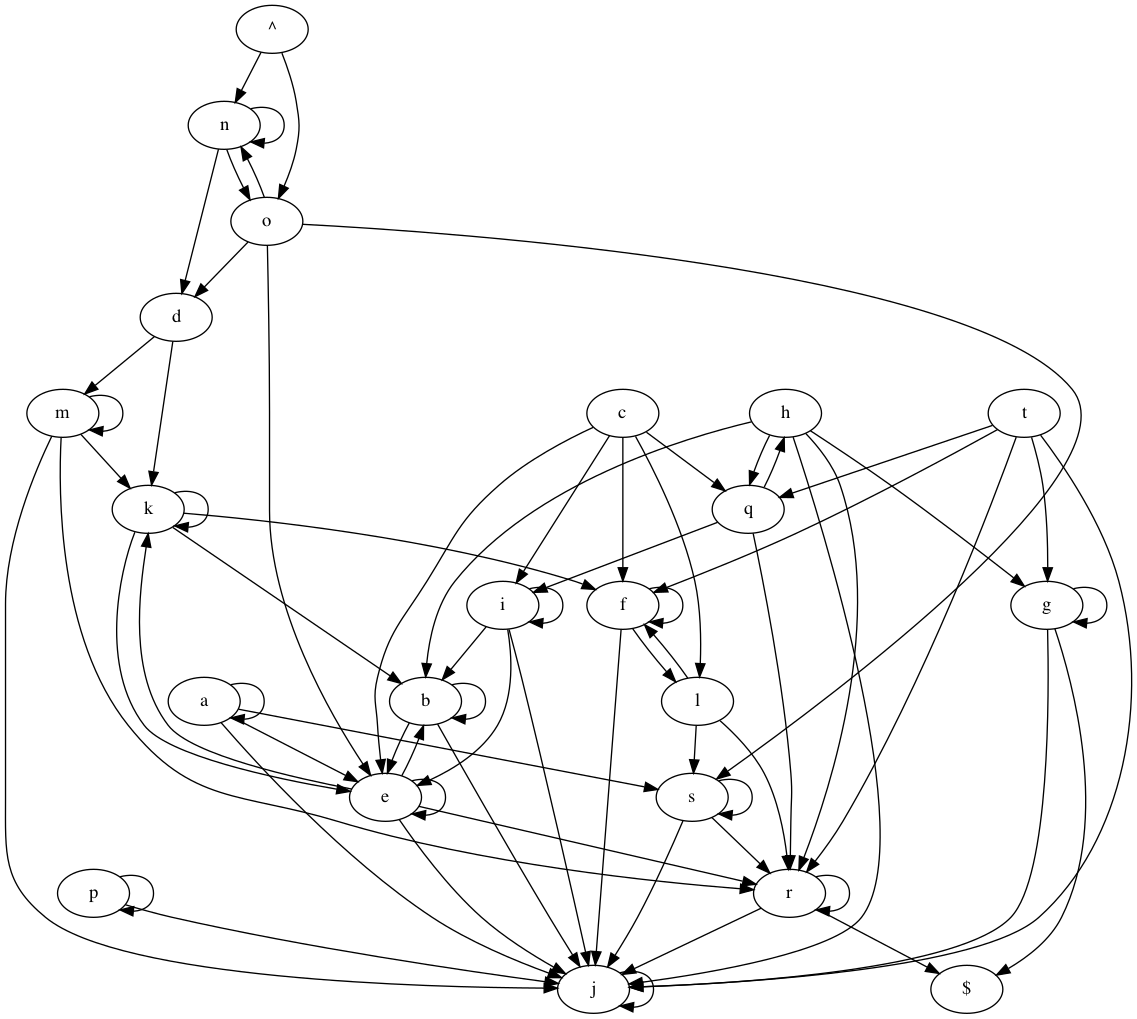}
\caption{Dialog-Structure}
\end{figure}

\paragraph{Dialog States}    
Those clusters were then used to map transition probabilities, all transitions with probability less than 0.6 were ignored. In Figure 1, the clusters are represented by alphabets a to t, start and end are represented as by \textasciicircum and \$ symbols. 
\newline
As we can see in figure 1, the dialog starts from \textasciicircum to n or o. n and o are self-introduction (I am x)  and rpc-inquiry (am I talking to y). From n and o it goes to other nodes including d which is brand-intro (I am calling from brand z). There are nodes like e, j, r which are common utterances like e (the reason why I called ..), j (please hold for a minute) and r (your transaction id is) which can be reached from most of the other nodes. r and g (when will it be done) are the most common termination nodes. Given below in the table are few of the node names along with the self labels. For our understanding a manual label column has also been added. These self labels are utterances from the actual dialogs which were mined using the self-labelling method of SCAN.

\begin{tabular}{ l | l | l}
  \hline
  cluster-name & manual-label & self-label \\
  \hline
  \textasciicircum & start-node & Start Node \\
  n & self-introduction & I am X  \\
  o & rpc-inquiry & Am I talking to Y  \\
  d & brand-intro & I am calling from brand Z \\
  m & product-info & This call is about product Z that you purchased last month \\
  k & payment-inquiry & Have you made the payment for the last installment \\
  f & amount-info & The amount is five thousand three hundred dollars \\
  l & date-reminder & Your due date is third of June \\
  r & number-intimation & Your transaction id is five nine eight zero double two   \\
  g & payment-date-inquiry & When will the payment be done   \\
  \$ & end-node & End Node
\end{tabular}

\bibliographystyle{unsrtnat}
\bibliography{references} 

\begin{thebibliography}{1}
	\bibitem{vangansbeke2020scan}
    \newblock Van Gansbeke, Wouter and Vandenhende, Simon and Georgoulis, Stamatios and Proesmans, Marc and Van Gool, Luc. Scan: Learning to classify images without labels. 
    \newblock In: {\em Proceedings of the European Conference on Computer Vision} (2020).
	


    \bibitem{qiu-etal-2020-structured}
	\newblock Qiu, Liang  and Zhao, Yizhou  and Shi, Weiyan  and Liang, Yuan  and      Shi, Feng  and Yuan, Tao  and Yu, Zhou  and Zhu, Song-Chun. Structured Attention for Unsupervised Dialogue Structure Induction.
	\newblock In :{\em Proceedings of the 2020 Conference on Empirical Methods in Natural Language Processing (EMNLP)}, pp. 1889--1899.  (2020).

	\bibitem{xu2020DDG}
	\newblock Jun Xu, Zeyang Lei, Haifeng Wang, Zheng-Yu Niu, Hua Wu, Wanxiang Che, Ting Liu. Discovering Dialog Structure Graph for Open-Domain Dialog Generation.
	\newblock arXiv preprint {\em arXiv:2012.15543}, (2020).

\end{thebibliography}

\end{document}